\documentclass[11pt,a4paper]{article}
\usepackage[hyperref]{ranlp2023}
\usepackage{times}
\usepackage{latexsym}

\usepackage{graphicx}
\usepackage{microtype}
\usepackage{float}
\usepackage{xcolor}
\aclfinalcopy

\title{Transformer-Based Model for Multilingual Hope Speech Detection}
\author{
  \textbf{Nsrin Ashraf\textsuperscript{1,2}},
  \textbf{Mariam Labib\textsuperscript{2,3}},
  \textbf{Hamada Nayel\textsuperscript{1,4}}
\\
  \textsuperscript{1}Department of Computer Science, Faculty of Computers and Artificial Intelligence,\\ Benha University, Egypt\\
  \textsuperscript{2}Computer Engineering, Elsewedy University of Technology, Cairo, Egypt\\
  \textsuperscript{3}Department of Electronics and Communications Engineering, Faculty of Engineering,\\ Mansoura University, Egypt\\ 
  \textsuperscript{4}Department of Computer Engineering and Information, College of Engineering,\\ Wadi Ad Dwaser, Prince Sattam Bin Abdulaziz University, Al-Kharj 16273, Saudi Arabia\\
  \small{
    \textbf{Correspondence:} \href{mailto:hamada.ali@fci.bu.edu.eg}{hamada.ali@fci.bu.edu.eg}
  }
}

\date{}

\begin{document}
\maketitle
\begin{abstract}
This paper describes a system that has been submitted to the "PolyHope-M" at RANLP2025. In this work various transformers have been implemented and evaluated for hope speech detection for English and Germany. RoBERTa has been implemented for English, while the multilingual model XLM-RoBERTa has been implemented for both English and German languages. The proposed system using RoBERTa reported a weighted f1-score of 0.818 and an accuracy of 81.8\% for English. On the other hand, XLM-RoBERTa achieved a weighted f1-score of 0.786 and an accuracy of 78.5\%. These results reflects the importance of improvement of pre-trained large language models and how these models enhancing the performance of different natural language processing tasks.       
\end{abstract}

\section{Introduction}  
In this era, and due to the tremendous usage of social media platforms, it is essential to automatically analyse their contents \citep{ashraf-etal-2025-reglat}. Hope speech detection has emerged as a critical  task for promoting healthy online discourse, with applications in mental health support and toxic content moderation. Unlike general sentiment analysis, hope speech emphasizes the identification of future-oriented language that fosters resilience and inclusivity \citep{chakravarthi-muralidaran-2021-findings}. The PolyHope-M shared task presented at IberLEF 2024 formalized this endeavor by establishing the inaugural multilingual benchmark for hope speech detection across English, Spanish, German, and Italian \citep{garcia2024overview}. Computational linguistics approach, using computational methods as well as linguistic features, reported high performance in various tasks such as sentiment analysis, hate speech detection, offensive language detection and sarcasm detection \citep{benha,labib-etal-2025-reglat, AbuElAtta2023}.     
\par Different approaches have been implemented to perform the aforementioned tasks, such as unsupervised learning, classical machine learning, and deep learning approaches. In the era of pre-trained Language Models (LMs), transformer-based models such as RoBERTa\footnote{\url{https://huggingface.co/docs/transformers/en/model_doc/roberta}} and XLM-RoBERTa\footnote{\url{https://huggingface.co/docs/transformers/en/model_doc/xlm-roberta}} have demonstrated efficacy in text classification tasks, their performance in the context of hope speech especially in non-English languages—remains insufficiently explored \citep{divakaran2024hope}.\\ 
\par The German language presents distinct challenges due to its morphological complexity, cultural differences in the expression of hope, and a scarcity of annotated datasets \citep{chan-etal-2020-germans}.\\
\par This paper seeks to address these gaps deficiencies by:
\begin{enumerate}
    \item Assessing the performance of RoBERTa for English and XLM-RoBERTa for German on the PolyHope-M dataset, comparing their performance against competition baselines.
    \item Developing a language-specific preprocessing pipeline to mitigate social media noise and to accommodate the intricacies of German orthography. 
\end{enumerate}

The rest of the paper paper is organized as follows: Section 2 reviews related work in hope speech detection. Section 3 describes the methodology, including dataset, preprocessing, and model architecture. Section 4 presents experiments results and discussion. Finally, section 5 concludes the results.
\section{Related Work}
Recent investigations into the detection of hope speech have examined a variety of methodologies across different languages. \citet{roy2022iiitsurat} established the effectiveness of traditional machine learning techniques, such as Support Vector Machines (SVMs) and logistic regression, for classifying hope speech in English, achieving a commendable accuracy of 78.2\% without the need for complex architectures. Following this, \citet{das2023hate} demonstrated that transformer-based models, including BERT and RoBERTa, enhanced performance by 9-12\% in terms of f1-score, particularly in the context of code-mixed social media text.

\par The exploration of multilingual approaches has gained momentum, as evidenced by \citet{yigezu2023multilingual}, who conducted a comparative analysis across 11 languages and found that the XLM-R model outperformed monolingual models in low-resource environments by an average of 6.3\% in f1-score. In the context of Spanish English interactions, \citet{divakaran2024hope} achieved state-of-the-art results with an f1-score of 84.1\% through the use of ensemble learning combined with cross-lingual embeddings. Additionally, \citet{arunadevi2024machine} emphasized the efficacy of hybrid CNN-LSTM architectures in addressing imbalanced datasets related to hope speech.

\par The practical implementation of these findings was furthered by \citet{aggarwal2023hope}, who incorporated hope speech detection into social media moderation frameworks, highlighting significant challenges associated with the real-time processing of colloquial expressions. Collectively, these studies illustrate the evolution of the field from basic classifiers to advanced multilingual systems, while also identifying ongoing challenges in managing morphologically rich languages and the complexities of noisy user-generated content.
\par The PolyHope-M shared task was introduced to enhance the computational modeling of hope by focusing on its identification and categorization in social media texts \citep{garcia2024overview}. This study aims to evaluate the effectiveness of RoBERTa-based model for English language and XLM-RoBERTa-based model for German language in multilingual text classification.
\section{Methodology}
\begin{figure*}[h]
    \begin{center}
    \includegraphics[width=\textwidth]{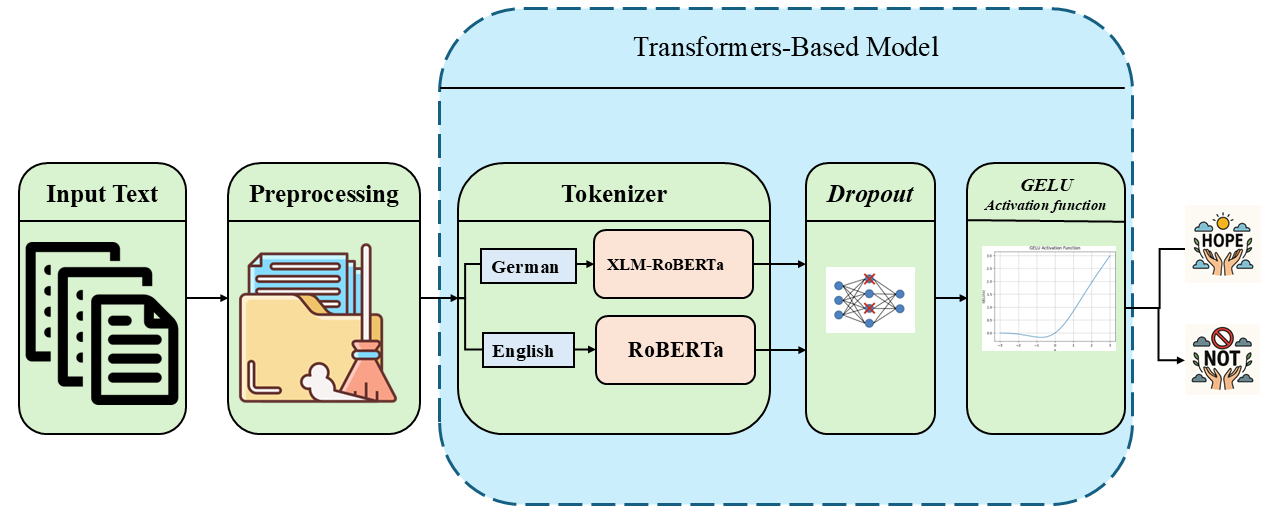}
    \caption{Overall Architecture of the Proposed Model}
    \label{fig:overall}
    \end{center}
\end{figure*}

The proposed approach follows a standard transformer-based pipeline for multilingual text classification, leveraging the strengths of pre-trained language models. The methodology pipeline consists of five key stages as shown in \ref{fig:overall}. It begins with input text, which is passed through the following pre-processing steps. The pre-processed text was passed through the RoBERTa tokenizer for English text to convert into sub-word tokens and corresponding input IDs. \par For German data, we used the XLM-RoBERTa tokenizer for German to leverage its multilingual capabilities. The tokenizers convert the input text into sub-word tokens and generate corresponding input IDs and attention masks that serve as inputs for the transformer models. The tokenized output then passes through a text embedding layer to create dense vector representations, capturing the semantic meaning of the input text. To prevent overfitting, a dropout layer is applied to the embeddings before passing them to the final classification layer. The classification layer is a fully connected neural network that transforms the text embeddings into output class probabilities using \texttt{GELU} activation function.
\subsection{Dataset }
The PolyHope dataset \citep{sidorov2023regret} consists of manually annotated social media texts in four languages. For shared task, English and German subsets of the dataset have been utilized to explore hope speech detection in multilingual settings. Each text instance in the dataset is labeled for {\bf Hope} and {\bf Not Hope}. A general statistics of class distribution over the dataset for English and German are shown in Table \ref{tab:polyhope_en_stats} and Table \ref{tab:polyhope_de_stats} respectively.\\
\begin{table}[H]
\centering
\begin{tabular}{|l|c|c|c|}
\hline
\textbf{Dataset} & \textbf{Size} & \textbf{Hope} & \textbf{Not Hope} \\
\hline
Train & 4541 & 2296 & 2245 \\
Dev   & 1650 & 834  & 816 \\
Test  & 2065 & --   & -- \\
\hline
\end{tabular}
\caption{Statistics of the PolyHope English dataset.}
\label{tab:polyhope_en_stats}
\end{table}
\begin{table}[H]
\centering
\begin{tabular}{|l|c|c|c|}
\hline
\textbf{Dataset} & \textbf{Size} & \textbf{Hope} & \textbf{Not Hope} \\
\hline
Train & 11573 & 4924 & 6649 \\
Dev   & 4208  & 1790 & 2418 \\
Test  & 5262  & --   & -- \\
\hline
\end{tabular}
\caption{Statistics of the PolyHope German dataset.}
\label{tab:polyhope_de_stats}
\end{table}
\subsection{Data preprocessing} 
A text cleaning technique using regular expression, implemented by \texttt{Regex} and \texttt{NLTK} libraries, was applied. Preprocessing comprises of the following steps:
\begin{itemize}
  \item Convert all characters in the text to lowercase. 
  \item Remove URLs.
  \item Punctuation, Symbol and Emojis Removal
  \item Punctuation and Symbol Removal (German-compatible), remove all characters except letters (including German umlauts: ä, ö, ü, ß, and their uppercase forms) and whitespace.
\end{itemize}
\subsection{Experimental Setup}
Various hyperparameters affect the performance of transformer models, such as the activation function, the number of hidden layers, the attention heads, the batch size, and the learning rate. As a large variety of hyperparameters can be used, we conducted many experiments to choose the optimal hyperparameters for the proposed model. Table \ref{tab:hyperparameters_comparison} shows the best hyperparameter values of the experimented transformer models that achieved the best results. 
\begin{table}[h]
\centering
\resizebox{\columnwidth}{!}{  
\begin{tabular}{|l|c|c|c|}
\hline
\textbf{Hyperparameter} & \textbf{German (XLM-R)} & \textbf{English (RoBERTa)} & \textbf{German (RoBERTa)} \\
\hline
Model Architecture & XLM-RoBERTa & RoBERTa-Base & RoBERTa-Large \\
Hidden Size & 768 & 768 & 768 \\
Number of Layers & 12 & 12 & 14 \\
Number of Attention Heads & 12 & 12 & 16 \\
Intermediate Size & 3072 & 3072 & 3072 \\
Dropout Rate & 0.2 & 0.1 & 0.1 \\
Activation Function & RELU & GELU & GELU \\
Max Sequence Length & 514 & 514 & 514 \\
Batch Size & 16 & 16 & 8 \\
Learning Rate & $1e^{-5}$ & $2e^{-5}$ & $4e^{-5}$ \\
Optimizer & AdamW & AdamW & RMSProp \\
Loss Function & CrossEntropyLoss & CrossEntropyLoss & Binary Cross-Entropy \\
\hline
\end{tabular}
}
\caption{Hyperparameters and model settings for German and English datasets using XLM-RoBERTa and RoBERTa.}
\label{tab:hyperparameters_comparison}
\end{table}
Performance evaluation of the submitted systems has been carried out using recall, precision, and f1-score \citep{9781707}. While, ranking the results was carried out using macro-averaged f1-score which is calculated by taking the unweighted mean of the per-class f1-scores.
\section{Results and discussion}
The experimental results highlight the effectiveness of transformer-based models for nuanced hope speech classification across languages. On the English dataset, RoBERTa demonstrated solid performance, especially considering the competitive nature of the track. The model ranked 14th out of 16 participants, achieved a macro-averaged f1-score of 0.818 with an overall accuracy of 81.84\%. 
\par For the German dataset, XLM-RoBERTa achieved a competitive performance in the German binary text classification. The model obtained a macro-avearged f1-score of 0.786 with an overall accuracy of 78.5\%. Among 12 participating teams, our model was ranked 12th, suggesting improved balancing performance across classes. The detailed results are shown in Table \ref{tab:_results}
\begin{table}[H]
\centering
\begin{tabular}{|c|c|c|}
\hline
\textbf{Metric} & \textbf{English} & \textbf{German} \\
\hline
Macro-averaged precision & 0.8199 & 0.7921 \\
\hline
Macro-averaged recall    & 0.8188 & 0.7973 \\
\hline
macro-averaged f1-score    & 0.8183 & 0.7851 \\
\hline
Accuracy        & 0.8184 & 0.7854 \\
\hline
\end{tabular}
\caption{PolyHope-M Test-phase Results for English and German datasets.}
\label{tab:_results}
\end{table}
\par An overview of the model's performance across binary classes and datasets is provided by the macro-averaged f1-score. The macro-averaged f1-score was 0.7851 for German and 0.8183 for English. A decrease in the macro-averaged f1-score for German is in line with previous results, which indicate that the model appears to have a harder time processing German text. While the model gets promising results in English dataset, in the German test the differences in language structure might not fully capture the syntactic and semantic differences, leading to reduced performance. The analysis indicates that the model is effective at identifying hope expressions, it performs better on the English dataset other than the German dataset. The differences in language structure and dataset size suggest areas for future work to get a better understanding models.
\section{Conclusion and Future work}
In this study, we proposed a transformer-based models for multilingual hope speech detection, focusing on English and German languages organized by PolyHope shared task. Our approach leveraged RoBERTa for English and XLM-RoBERTa for German, achieving competitive results. While the models achieved strong performance in English, the lower scores in German suggest challenges in handling language-specific nuances and morphologies, possibly due to differences in linguistic structure or dataset imbalance.
\par For future research, we plan to enhance model adaptability across languages through language-specific fine-tuning techniques and the use of more robust multilingual architectures. Although, testing larger multilingual models (e.g., mBERT, DeBERTa) or hybrid approaches combining transformers with linguistic features. Investigating techniques to avoid imbalance data and oversampling.


\bibliographystyle{acl_natbib}
\bibliography{ranlp2023}


\end{document}